\documentclass[final]{cvpr}

\usepackage{times}
\usepackage{epsfig}
\usepackage{graphicx}
\usepackage{amsmath}
\usepackage{amssymb}

\usepackage{caption}

\usepackage[switch]{lineno}
\usepackage[ruled,vlined]{algorithm2e}
\usepackage{mathtools}
\usepackage[flushleft]{threeparttable}
\usepackage{booktabs}
\usepackage{enumerate}
\usepackage{enumitem}
\usepackage{color}
\usepackage{xspace}
\newcommand{\Olala}{\texttt{OLALA}\xspace}

\usepackage[pagebackref=true,breaklinks=true,colorlinks,bookmarks=false]{hyperref}

\def\cvprPaperID{3373} % *** Enter the CVPR Paper ID here
\def\confYear{CVPR 2021}

\begin{document}

%%%%%%%%% TITLE
\title{\Olala: Object-Level Active Learning for Efficient Document Layout Annotation}

\author{Zejiang Shen\\
Harvard University\\
\and
Jian Zhao\\
University of Waterloo\\
\and 
Melissa Dell\\
Harvard University\\
\and
Yaoliang Yu\\
University of Waterloo\\
\and 
Weining Li\\
University of Waterloo\\
}

\maketitle

%%%%%%%%% ABSTRACT
\begin{abstract}

Document images often have intricate layout structures, with numerous content regions (\eg texts, figures, tables) densely arranged on each page.
This makes the manual annotation of layout datasets expensive and inefficient. 
These characteristics also challenge existing active learning methods, as \textit{image-level} scoring and selection suffer from the overexposure of common \textit{objects}.
Inspired by recent progresses in semi-supervised learning and self-training, we propose an \textbf{O}bject-\textbf{L}evel \textbf{A}ctive \textbf{L}earning framework for efficient document layout \textbf{A}nnotation, \Olala. 
In this framework, only regions with the most ambiguous object predictions within an image are selected for annotators to label, optimizing the use of the annotation budget. 
For unselected predictions, the semi-automatic correction algorithm is proposed to identify certain errors based on prior knowledge of layout structures and rectifies them with minor supervision. 
Additionally, we carefully design a perturbation-based object scoring function for document images. 
It governs the object selection process via evaluating prediction ambiguities, and considers both the positions and categories of predicted layout objects.  
Extensive experiments show that \Olala can significantly boost model performance and improve annotation efficiency, given the same labeling budget.  
Code for this paper can be accessed via \url{https://github.com/lolipopshock/detectron2_al}. 

\end{abstract}

\section{Introduction}

\begin{figure}[t]
    \centering
    \includegraphics[width=0.9\linewidth]{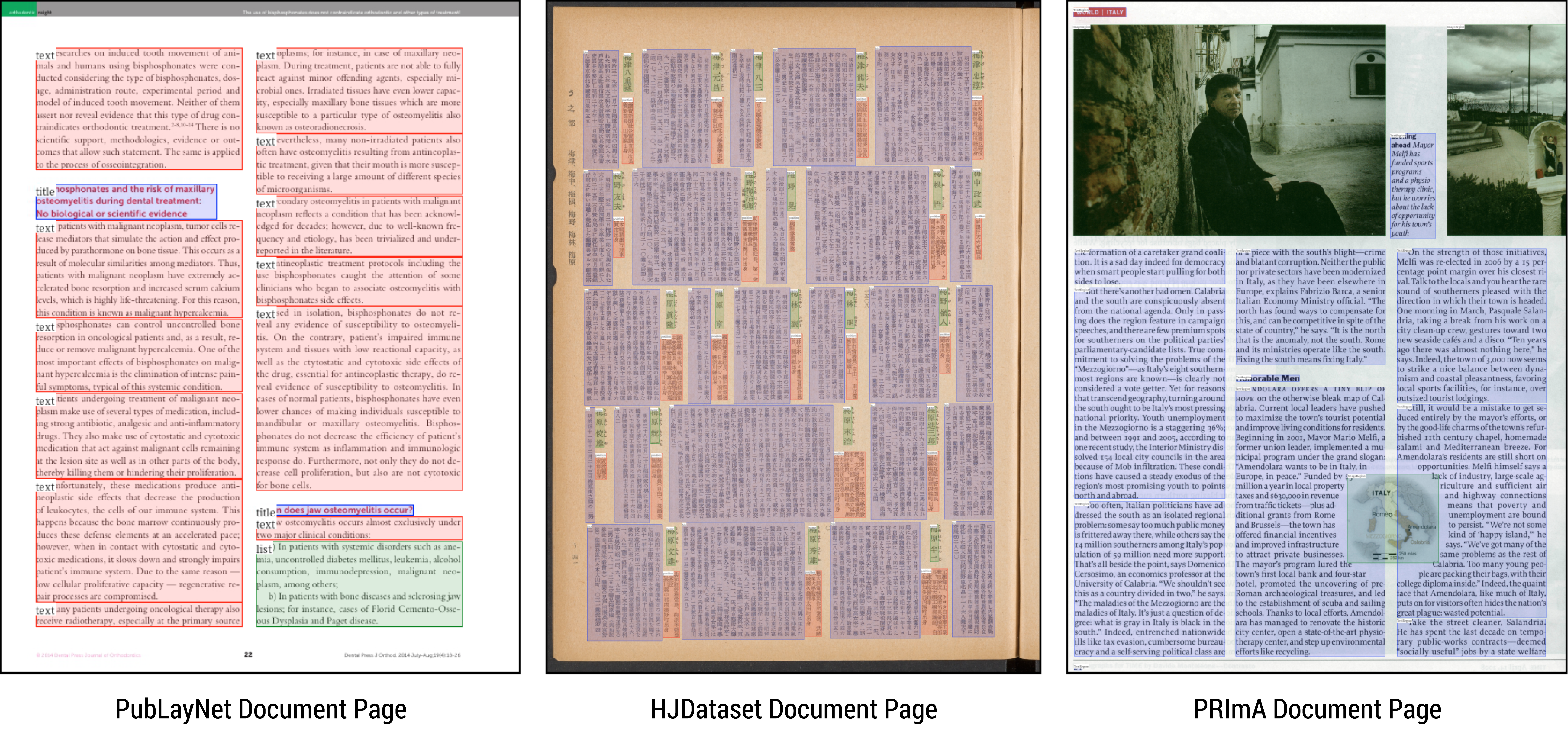}
    \caption{Three exemplar document layouts from PublayNet~\cite{zhong2019publaynet}, HJDataset~\cite{shen2020large}, and PRImA~\cite{antonacopoulos2009realistic}. There are numerous layout objects per page, many of them are very similar and from the same category. Directly labeling them all will result in waste of precious labeling budget.} 
    \label{fig:example}
    \vspace{-2mm}
  \end{figure}
\begin{figure*}[t]
    \centering
    \includegraphics[width=\linewidth]{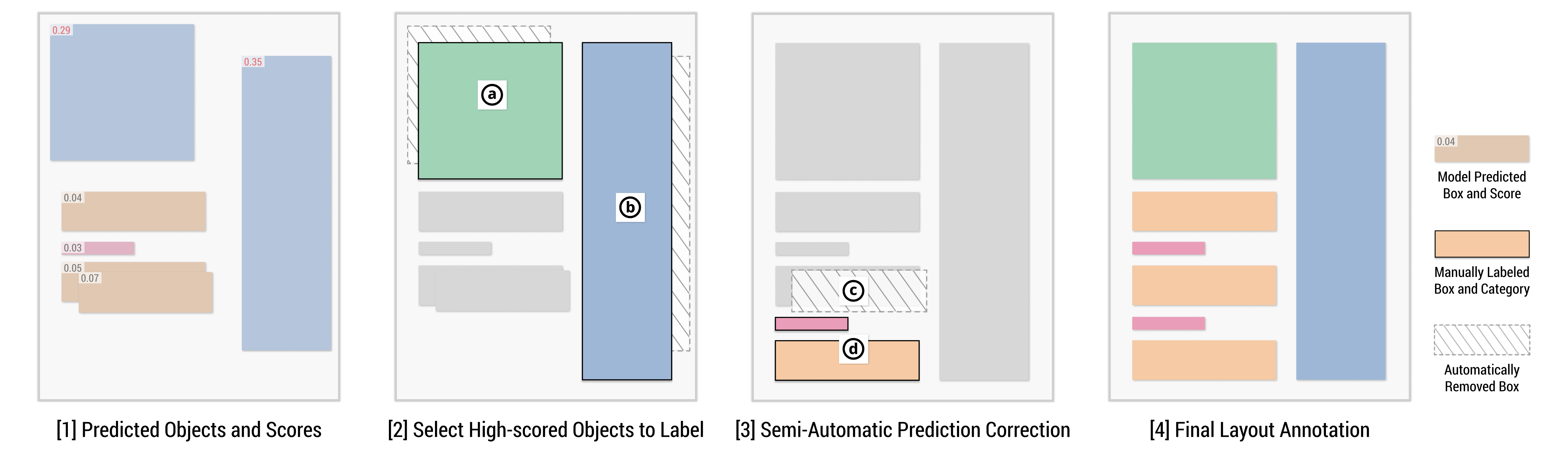}
    \caption{Illustration of the \Olala framework. 
    [1] During labeling, for an input image, a trained model predicts the layout with various errors. A object scoring function $f$ evaluates the informativeness for each object prediction. 
    [2] \Olala selects the regions of top scores and sends them for manual labeling to correct wrong object category (a) and bounding box (b). 
    [3] A semi-automatic prediction correction algorithm is applied to rectify duplicated object (c) and recover false-negatives (d) with minimal extra supervision. 
    [4] After this process, the final annotation is obtained with labeling only a part of the objects.} 
    \label{fig:method}
    \vspace{-3mm}
\end{figure*}

Deep learning-based approaches have been widely applied to document layout analysis and content parsing~\cite{zhong2019publaynet, schreiber2017deepdesrt}.
Illustrated in Figure~\ref{fig:example}, document layout object detection, like image object detection, requires identifying content regions and categories within images. 
A key distinction, however, is that it is common for dozens to hundreds of content regions to appear on a single page in documents (\eg, firm financial reports or newspapers), as opposed to only several objects per image in natural image datasets (\eg,  5 in the MS-COCO Dataset~\cite{lin2014microsoft}). 
Additionally, the region category distribution is heavily imbalanced and requires more pages to be annotated to allow for reasonable exposure of uncommon categories (\eg, footnotes and watermarks). 
Hence, the manual labeling process often used on natural images to create high-quality labeled datasets can be prohibitively costly to replicate for documents of central interest to academic researchers and business organizations. 

Recent studies have shown that document images contain certain layout structures that can help reduce labeling cost. 
For example, PubLayNet~\cite{zhong2019publaynet} and TableBank~\cite{li2019tablebank} Dataset are created via rule-based parsing of electronic PDF documents or source files.
However, solely relying on rules restricts the layout diversity and complexity in the datasets, and is difficult to generalize to scanned documents. 
Most recently, HJDataset~\cite{shen2020large} is created via manually modifying layouts generated by a layout prediction model for scanned historical archives. 
Yet the model is designed ad-hoc for the dataset, calling for generalizable approaches for incorporating prior knowledge about layouts in the labeling process.  

Active Learning (AL) has been widely adopted in image object detection for optimizing labeling efficiency via prioritizing the most important samples to annotate~\cite{aghdam2019active, haussmann2020scalable, brust2018active, roy2018deep}.
However, while the end goal is to annotate individual objects within an image, these AL methods typically score and select samples at the image level, rather than at the object level. 
For category-imbalanced layout images, it could suffer from the over-exposure of common objects, leading to suboptimal results. 
Recent advances in Semi-Supervised Learning (SSL) and self-training can boost model performance using unlabeled data~\cite{rosenberg2005semi, xie2020self}. 
The Self-supervised Sample Mining (SSM) algorithm~\cite{wang2018towards} proposes to stitch high-confidence patches from unlabeled data to labeled data to improve both the labeling efficiency and model performance. 
It enables object-level prediction selection and has the potential to over sample less common layout objects. 
However, their approach requires objects being sparsely distributed per image, making it not applicable to our case where content regions are densely arranged on document images. 

To address these challenges, we propose a novel AL framework, \Olala, an \textbf{O}bject-\textbf{L}evel \textbf{A}ctive \textbf{L}earning framework for efficient layout \textbf{A}nnotation. 
Shown in Figure~\ref{fig:method}, in this framework, critical \emph{objects}, rather than \emph{images}, are individually evaluated and selected for labeling.
During the labeling process, \Olala trains a model to generate object predictions. 
Within an image, only regions of most ambiguous predictions are chosen for human inspection and annotation, addressing the inefficient use of annotation budget on common objects or categories.
Central to this process is a semi-automatic prediction correction algorithm. 
Inspired by the previous endeavors of automated layout dataset generation~\cite{zhong2019publaynet, li2019tablebank, shen2020large}, it attempts to incorporate prior knowledge about layout structures to ensure the high quality of the created dataset. 
It can identify false-positives and false-negatives in the unselected model predictions, and correct them with minimal extra supervision. 
Additionally, we design a novel object-level scoring function governing the region selection process. 
The perturbation-based scoring method evaluates consistency of both object position and category predictions between the original and perturbed inputs. 
Compared to prior work, it is carefully designed for layout datasets with unique arrangement of content regions, and can identify errors of critical importance to layout analysis tasks.  
To the best of our knowledge, this is the first AL method dedicated to document layout analysis. 

Through extensive experiments, we study how the proposed approach can improve labeling efficiency in two different scenarios. 
Given sufficient labeling budgets, we show that \Olala can create datasets with better trained model performance compared to image-level AL baselines. 
This is a typical evaluation for AL methods; by conducting a fine-grained object-level selection, our approach can improve the model accuracy by a huge margin. 
On the other side, as only part of an image requires annotation in our method, we demonstrate that our method can create datasets of the same size with far less human effort. 
It is extremely helpful when a tight labeling budget is available, or the labeling task is time-sensitive. 

Our main contributions can be summarized as follows:
\setlist{nolistsep}
\begin{itemize}[noitemsep]
    \item We introduce \Olala, a framework that actively selects the most important \emph{objects}, rather than images, for efficient annotation of document layouts. Possible errors in unselected object predictions can be identified using a semi-automatic prediction correction algorithm (Section~\ref{sec:method}). 
    \item We propose a perturbation-based object scoring method that selects the most ambiguous objects for annotation. It considers both object position and category, directly applicable to object-dense layout images (Section~\ref{sec:scoring}).
    \item We demonstrate the practical use of \Olala for improving labeling efficiency and model performance under different scenarios (Section~\ref{sec:exp} and \ref{sec:res}). 
\end{itemize}

\section{The \Olala Framework}
\label{sec:method}

\subsection{Object-Level Active Learning Setup}
In layout object detection problems (see Section \ref{sec:related} for an overview), a detection model $\Theta$ is trained to identify $n_i$ objects within an input image $X_i$, where the bounding box $b_j$ and category distribution $c_j$ is estimated for the $j$-th object. 
$Y_i=\{(b_j,c_j)\}^{n_i}_{j=1}$ are the object annotations for $X_i$. 
$\Theta$ is initially trained on a small labeled dataset $\mathcal{L}_0=\{(X_i,Y_i)\}_{i=1}^l$, and it receives a large unlabeled dataset $\mathcal{U}_0=\{X_i\}_{i=1+l}^{u+l}$. 

The goal of typical image-level AL methods~\cite{aghdam2019active, brust2018active, roy2018deep} is to optimally sample images from $\mathcal{U}$ for annotation to maximally improve the model's performance on given metrics. 
This process could be iterative: at each round $t$, it selects $m$ samples $\mathcal{M}_t=\{X_i\}_{i=1}^m$ from $\mathcal{U}_{t-1}$ to query labels, obtains the corresponding labeled set $\bar{\mathcal{M}_t}=\{(X_i, Y_i)\}_{i=1}^m$, and updates the existing labeled set $\mathcal{L}_t=\mathcal{L}_{t-1}\cup \bar{\mathcal{M}_t}$.
The new model $\Theta_t$ is obtained by training (or fine-tuning) on $\mathcal{L}_t$. 
For the next round, the unlabeled set becomes $\mathcal{U}_t=\mathcal{U}_{t-1} \setminus \mathcal{M}_t$. 
In this process, annotators need to create all objects labels $\bar{Y_i}=Y_i$ for the images in $\mathcal{M}_t$.
This is not optimal for layout object detection, where many objects could appear on a single image. 
Because of the uneven distribution of objects, sometimes only a small portion of object predictions in an image are inaccurate.
Labeling whole images wastes budget on these accurately predicted regions, which could be otherwise used for labeling less accurate objects.

Consider an alternative setup illustrated in Figure~\ref{fig:method}: the AL agent prioritizes annotation for a portion of objects in $Y_i$ within each image. 
An \emph{object}-level scoring function $f$ evaluates the ambiguities of predictions generated by $\Theta$. 
Object regions of top scores, the \emph{selected objects}, will be sent for manual annotation to create labels $\bar{Y_i}$. 
To wisely use human efforts, the ratio of selected objects $r$ is dynamically adjusted during the labeling process (Section \ref{sec:scheduling}). 
And after correcting possible errors (Section \ref{sec:correction}), the remaining \emph{unselected objects} constitute the complement labels $\hat{Y_i}$ and are merged with the human labels.
The \emph{Objects Selection Scheduling} and \emph{Semi-automatic Prediction Correction} ensures the combined annotation $\tilde{Y_i}=\bar{Y_i}\cup\hat{Y_i}$ can be approximately close to $Y_i$. 
Therefore, accurate dataset annotations can be created with only $|\bar{Y_i}|/|\tilde{Y_i}|$ of time  ($|\cdot|$ being the cardinality of the set), and more images can be annotated given the same labeling budget.  
This is our object-centered labeling setup in \Olala. 

\subsection{Objects Selection Scheduling} 
\label{sec:scheduling}

The ratio of selected objects during training can both influence the labeling efficiency as well as the trained model accuracy.
A ratio near $1$ approximates the whole human labeling process (less efficient), while zero ratio resembles a full self-training~\cite{rosenberg2005semi} settings (less accurate). 
To optimally balance the efficiency and accuracy, $r$ is dynamically adjusted at different rounds of labeling via a scheduling function. 
According to Curriculum Learning~\cite{bengio2009curriculum}, we set high initial values of $r$ to rely more on human labeling as an attempt to ease the model training in the beginning. 
Linear or exponential decay is then applied to gradually decrease $r$, and we increase the trust in the model predictions as their accuracy improves during training. 
From an optimization perspective, $r$ can be seen as a ``learning rate'' for the \Olala AL process. 
Appropriate $r$ settings can help correct most false-positives in model predictions (the model generates the wrong bounding box or class for an object). 
And we demonstrate the effectiveness of the proposed scheduling mechanism in the experiments (Section~\ref{sec:exp-olala}). 

\subsection{Semi-automatic Prediction Correction}
\label{sec:correction}

Compared to recent work~\cite{wang2018towards, xie2020self} using self-training for improving model performance (see Section~\ref{sec:related}), \Olala consists an additional component to fix possible errors in the used model predictions.
Inspired by recent efforts for creating large-scale layout analysis datasets~\cite{zhong2019publaynet, li2019tablebank, shen2020large}, we propose a semi-automatic prediction correction algorithm to ensure the quality of the model predictions. 
This method relies on the unique structures of document data: layout objects are densely arranged, and there is usually no overlapping between content regions. 
It can identify \emph{duplicated prediction} and \emph{false-negative predictions} based on this prior knowledge, and requests minor supervision to fix them. 
Shown in Section~\ref{sec:exp-same-object} and \ref{sec:exp-same-image}, this algorithm both improves the final trained model accuracy, and enables the creation of an accurate large dataset based on these predictions\footnote{Self-training methods (\eg \cite{wang2018towards}), usually discard the model predictions (pseudo labels) after training.}.

\textbf{Duplication Removal} 
In practice, models could generate multiple close predictions for a large object, yet only one or some of the predictions are sent for user inspection. 
Thus, if naively merging the user's labels with the remaining predictions, it can lead to overlapping labels for the same object. 
This is not compliant with the usual layout structures, and will mislead the trained model to produce overlapping boxes with high confidences. 
We fix this error by filtering out predictions overlapped with any human annotations over a score threshold $\xi$. 
Different from IOU scores, we use the the pairwise Overlap Coefficient, $\text{Overlap}(A,B) = {|A\cap B|}/{\min(|A|, |B|)}$, to better address scenarios where a predicted box is contained within a labeled box. 
The threshold $\xi$ is set to 0.25 empirically.  

\textbf{Missing Annotation Recovery} 
False-negatives occur when no prediction is generated for a given object. 
In typical object detection tasks, predictions are dropped when the confidence is under some threshold, which might lead to false negatives. 
It is an implicit signal from the model, requesting extra supervision from human annotators; we show that it is the key step for improving the dataset accuracy (Section~\ref{sec:exp}). 
It is implemented as highlighting the regions without model predictions, such that human annotators (or a simulated agent) can easily identify the mis-predicted objects and add the annotations. 

The implementation of this algorithm is different between real-world human annotation (without oracle beforehand) and simulated labeling experiments (with oracle beforehand). 
For human annotations, we carefully design a user interface which incorporates the three functions and augments human labeling, and we refer readers to the supplementary material for more details. 
In simulations, we build a labeling agent that can automatically query the oracle for ground-truths under different scenarios (see Section~\ref{sec:exp}). 

\subsection{Overview of the Proposed Algorithm}

We now present the formal \Olala Algorithm~\ref{alg:olala}. 
Given an initial labeled set $\mathcal{L}_0$, it aims to use the predictions from a model $\Theta$ to optimally label the remaining unlabeled set $\mathcal{U}_0$ given some labeling budget. 
Different from existing work, we define the labeling budget per round $m$ as the number of \emph{objects} rather images that human annotators can label. 
The algorithm iteratively proposes the most informative objects to label for a total of $T$ rounds. 
At each round $t$, it selects up to $m$ objects to label. 
For each image $X_i$ from the existing unlabeled set $\mathcal{U}$, $r$ percent of predicted objects are selected for user labeling according to some object scoring function $f$. 
The rest of the labels are created by correcting errors in the unselected model prediction $\hat{Y}_i^-$ based on the semi-automatic prediction correction algorithm. 
The labeled image $X_i$ will be removed from $\mathcal{U}$ and the annotated samples $(X_i, \tilde{Y_i})$ will be added to $\mathcal{L}$. 
After each round, the selection ratio $r$ decays as the model accuracy improves.

\begin{algorithm}[t]
\SetAlgoLined
\SetKwInOut{Input}{Input}
%\SetKwInOut{Initialize}{Initialize}
%\KwResult{Write here the result }
    \Input{Initial sets $\mathcal{U}_0$, $\mathcal{L}_0$; labeling budget $m$; object selection ratio $r$}
    Initialize $\mathcal{U}=\mathcal{U}_0$, $\mathcal{L}=\mathcal{L}_0$, and model weights $\Theta$; \\
    \For{$t=0$ \KwTo $T-1$}{
    Calculate budget $m$ and selection ratio $r$ for at $t$ \\
    Update the model $\Theta$ using $\mathcal{L}$\\
    Let $\bar{\mathcal{M}}$ = \{\} \\
    \For{$i=0$ \KwTo $|\mathcal{U}|$}{
        Generate object predictions $\hat{Y}_i$ for $X_i\in\mathcal{U}$ \\
        Let $m_i$ = $\min\{r|\hat{Y}_i|, m\}$, $m=m-m_i$ \\
        \lIf{$m \leq 0$}{break}
        Calculate object scores $f(\hat{y}_j)$ $\forall\hat{y}_j\in\hat{Y}_i$ \\
        Select $m_i$ objects of top scores and label $\bar{Y_i}$ \\ 
        Correct errors in unselected predictions $\hat{Y}_i^-$ \\
        Merge $\bar{Y_i}$ with $\hat{Y}_i$ for image annotations $\tilde{Y_i}$ \\
        Remove $X_i$ from $\mathcal{U}$ and add $(X_i, \tilde{Y_i})$ to  $\bar{\mathcal{M}}$ \\
    }
    Update $\mathcal{L}\leftarrow\mathcal{L}\cup\bar{\mathcal{M}}$
    }
    Update the model $\Theta$ using $\mathcal{L}$\\
    \caption{Object-level Active Learning Annotation}
    \label{alg:olala}
\end{algorithm}

\section{Perturbation-based Scoring Function}
\label{sec:scoring}
The scoring function $f$ also plays an important role in the \Olala framework. It evaluates prediction ambiguity and determines which objects to select for labeling. 
We propose a perturbation-based scoring method based on both the bounding box and category predictions to account for specific characteristics in layout detection tasks. 
Inspired by the self-diversity idea in~\cite{jiang2020camouflaged} and \cite{zhou2017fine}, the proposed method hypothesizes that the adjacent image patches share similar features vectors, and the predicted object boxes and categories for them should be consistent.
Therefore, any large disagreement between the original and perturbed predictions indicates that the model is insufficiently trained for this type of input, or there is some anomalies in the given sample. 
Both cases demand user attention, and extra labeling is required. 

Specifically, for each object prediction $\hat{y}_j=(\hat{b}_j, \hat{c}_j)\in \hat{Y_i}$, we take the bounding box prediction $\hat{b}_j=(x,y,w,h)$ and apply some small shifts to perturb the given box, where $x, y$ are the coordinate of the top left corner, and $w,h$ are the width and height of the box. 
The new boxes are created via horizontal and vertical translation by a ratio of $\alpha$ and $\beta$: $p_{jk}=(x\pm\alpha w, y\pm\beta h, w, h)$, where $p_{jk}$ is the $k$-th perturbed box for box prediction $\hat{b}_j$, and a total of $K$ perturbations will be generated. 
Based on the image features within each $p_{jk}$, the model generates new box and category predictions $(q_{jk}, v_{jk})$. 
We then measure the disagreement between the original prediction $(\hat{b}_j, \hat{c}_j)$ and the perturbed versions $\{(q_{jk}, v_{jk})\}_{k=1}^K$, and use it as a criterion for selecting objects for labeling. 

In practice, we build this method upon a typical object detection architecture composed of two stages~\cite{ren2015faster}: 1) a region proposal network estimates possible bounding boxes, and 2) a region classification and improvement network (ROIHeads\footnote{It's a module name in Detectron2 \cite{wu2019detectron2}.}) predicts the category and modifies the box prediction based on the input proposals. 
We use the perturbed boxes $\{p_{jk}\}_{k=1}^K$ as the new inputs for the ROIHeads, and obtain the new box and class predictions $\{(q_{jk}, v_{jk})\}_{k=1}^K$. 
For object regions of low confidence, the new predictions are unstable under such perturbation, and the predicted boxes and category distribution can change drastically from the original version. 
To this end, we formulate the position disagreement $D_p$ and the category disagreement $D_c$ for the $j$-th object prediction as
\begin{align*}
    D_p(\hat{b}_j) & = \frac{1}{K}\sum_k{\left(1-\text{IOU}(\hat{b}_j, p_{jk})\right)} \\
    D_c(\hat{c}_j) & = \frac{1}{K}\sum_k{L(\hat{c}_j||v_{jk})},
\end{align*}
where $\text{IOU}$ calculates the intersection over union scores for the inputs, and $L(\cdot || \cdot)$ is a measurement for distribution difference, e.g., cross entropy. 
The overall disagreement $D$ is defined as $D(\hat{y}_j)=D_p(\hat{b}_j)+\lambda D_c(\hat{c}_j)$, with $\lambda$ being a weighting constant. 
Objects of larger $D$ will be prioritized for labeling, and users will create annotations $\bar{Y_i}$ for them in the $i$-th image. 

The proposed method thoroughly evaluates the box and category prediction robustness, and can effectively identify false-positive object predictions. 
Based on the self-diversity assumption, incorrect category prediction $\hat{c}_j$ will cause high $D_c$ because of the divergence of the new class prediction $v_{jk}$ for nearby patches. 
When the predicted box $\hat{b}_j$ is wrong, the perturbed box $p_{jk}$ is less likely to be the appropriate proposal box. 
The generated predictions $(q_{jk}, v_{jk})$ are unreliable, causing higher overall disagreement $D$.

\textbf{Applicability to Layout Datasets} Compared to previous work, the perturbation-based scoring function aims to solve two challenges unique to layout analysis tasks. 
First, different from real-world images, layout regions are boundary-sensitive: a small vertical shift of a text region box could cause the complete disappearance of a row of texts. 
However, existing methods designed for image-level selection usually focus on the categorical---rather than positional---information in object detection model outputs (i.e. \cite{brust2018active}, which considers the marginal score of the object category predictions and does not use the bounding boxes, or \cite{aghdam2019active}, which indirectly uses the positional information based on a pixel map for image-level aggregation). 
By contrast, our method could identify samples that lead to ambiguous boundary predictions via $D_p$, which explicitly analyzes the box prediction quality.

As an additional challenge, document images usually contain numerous objects per page and content regions are densely arranged in certain structures. 
This makes it not applicable to adapt the object-level scoring function in \cite{wang2018towards}, which requires cropping an object and randomly pasting it to another image and evaluates the consistency between the original and the newly detected boxes for this object. 
The random pasting will introduce non-existing structures (\eg, overlaying a figure over tables or texts), and the calculated score cannot reliably assess the prediction. 
To the contrary, in our method, we do not change original document images but only perturb the box predictions $\hat{b}_j$. 
The original document structures are untouched, and the scores could accurately indicate the object prediction informativeness.

\begin{table}[t]
    \resizebox{1.\linewidth}{!}{
    \begin{threeparttable}
        \begin{tabular}{l|l|l|l}
        \toprule
        \textbf{Datasets}             & \textbf{PubLayNet} &\textbf{HJDataset} & \textbf{PRImA}  \\
        \midrule
        Data Source                   & Digital PDF        & Image Scan        & Image Scan      \\
        Annotation                    & Auto PDF Parsing   & Combined          & Human Labeling  \\
        Dataset Size                  & 360,000            & 2,048             & 453             \\
        Train~/~test split            & 8,896~/~2,249      & 1,433~/~307       & 363~/~90        \\
        Avg~/~max $O$                 & 10.72~/~59         & 73.48~/~98        & 21.63~/~79      \\
        \midrule
        Labeling budget $m$           & 21,140 (2,000)     & 51,436 (700)      & 5,623 (240)     \\
        Total rounds $T$              & 10                 & 8                 & 4               \\
        Initial~/~last $r$            & 0.9~/~0.4          & 0.9~/~0.5         & 0.9~/~0.75      \\
        \bottomrule
        \end{tabular}
    \end{threeparttable}
    }
\caption{Statistics and parameters for the PubLayNet, HJDatasets, and PRImA. $O$ is the number of objects in each image. For the labeling budget, the numbers in the parentheses indicate the equivalent numbers of images of the given object labeling budget.}
\label{table:dataset}
\vspace{-3mm}
\end{table}
% \begin{table}[]
%     \begin{tabular}{rcccccc}
%     Datasets          & \multicolumn{2}{c}{PubLayNet} & \multicolumn{2}{c}{HJData} & \multicolumn{2}{c}{PRImA} \\
%     Experiments       & AP             & Total I/O    & AP            & Total I/O  & AP            & Total I/O \\
%     Image-Random      & 60.73          & 2046/21430   & 69.82         & 709/51959  & 31.49         & 244/4799  \\
%     OLALA-Random      & 64.21+5.73\%   & 3187/21412   & 72.16+3.36\%  & 1105/51626 & 32.08+1.89\%  & 277/4785  \\
%     Image-Marginal    & 67.91          & 2465/21574   & 73.25         & 709/51937  & 30.99         & 243/4769  \\
%     OLALA-Marginal    & 69.23+1.93\%   & 3661/21467   & 71.48-2.42\%  & 1075/51804 & 32.85+6.00\%  & 306/4721  \\
%     OLALA-Pertubation & 69.13+1.79\%   & 3686/21430   & 73.40+0.20\%  & 1159/51656 & 33.87+9.30\%  & 286/4764 
%     \end{tabular}
% \end{table}

\begin{table*}[t]
    \centering
    \resizebox{0.9\linewidth}{!}{
    \begin{threeparttable}
        \begin{tabular}{r|cc|cc|cc}
        \toprule
        Datasets          & \multicolumn{2}{c|}{PubLayNet} & \multicolumn{2}{c|}{HJData} & \multicolumn{2}{c}{PRImA\tnote{1}} \\
        \midrule
        Experiments       & Final AP             & Labeled $I$/$O$    & Final AP            & Labeled $I$/$O$  & Final AP            & Labeled $I$/$O$ \\
        \midrule
        Image-Random {[}a{]}     & 60.73        & 2046/21430   & 69.82         & 709/51959  & 31.49         & 244/4799  \\
        \Olala-Random {[}c{]}      & \textbf{64.21(+3.48)}\tnote{2}     & 3187/21412   & \textbf{72.16(+2.34)}  & 1105/51626 & \textbf{32.08(+0.59)}  & 277/4785  \\
        \midrule
        Image-Marginal {[}b{]}    & 67.91         & 2465/21574   & 73.25         & 709/51937  & 30.99         & 243/4769  \\
        \Olala-Marginal {[}d{]}   & \textbf{69.23(+1.31)}\tnote{3}    & 3661/21467   & 71.48(-1.77)  & 1075/51804 & 32.85(+1.86)  & 306/4721  \\
        \Olala-Pertubation {[}e{]} & 69.13(+1.21)   & 3686/21430   & \textbf{73.40(+0.15)}  & 1159/51656 & \textbf{33.87(+2.88)}  & 286/4764  \\
        \bottomrule
    \end{tabular}
    \begin{tablenotes}
        \small
        \item[1] The results in PRImA are averaged from the 5-folds in cross validation to account for possible noise due to the small dataset size.
        \item[2,3] The \Olala-Random percentages are compared against Image-Random, and others are compared against Image-Marginal. 
    \end{tablenotes}
    \end{threeparttable}
    }
    \caption{The final AP and number of total labeled images $I$ and objects $O$ given the same object budget $m$. \Olala achieves strong performance improvements in model accuracy in all experiments, and creates datasets with considerably more images given the same labeling budget.}
    \label{table:performance-object-fixed}
    \vspace{-5mm}
\end{table*}

\section{Experimental Setup}
\label{sec:exp}

\textbf{Objective} 
Several experiments (labeling simulations) are designed to study the validity of the proposed \Olala framework and evaluate how it could improve the efficiency of the labeling process. 
We quantify labeling efficiency as the model accuracy against the used amount of budget. 
Methods are considered better if they achieve similar accuracy while use less labeling budget $m$ than their counterparts, or obtain better accuracy given the same $m$.
In the experiments, we measure object detection accuracy using mean Average Precision (AP) scores~\cite{lin2014microsoft}, and the labeling budget refers to the number of objects to label by default. 

\textbf{Datasets} 
To validate our approach, we run simulations on three representative layout analysis datasets: PubLayNet~\cite{zhong2019publaynet}, PRImA~\cite{antonacopoulos2009realistic}, and HJDataset~\cite{shen2020large}. 
PubLayNet is a large dataset of 360k images. 
The images and annotations are generated from noiseless digital PDF files of medical papers. 
As the original training set in PubLayNet is too large to conduct experiments efficiently, we use a downsampled version of 8996 and 2249 samples for training and validation, respectively. 
PRImA is created by human annotators drawing bounding boxes for text regions in both scanned magazines and technical articles, resulting in greater heterogeneity in this dataset than in PubLayNet. 
We convert the original dataset into COCO format, and divide into the training (363 images) and validation (90 images) sets. 
HJDataset contains layout annotation for 2k historical Japanese documents. 
It has an intermediate dataset size, and shares similar properties with both the aforementioned datasets. 
HJDataset is established using noisy image scans, and the creation method is a combination of rule-based layout parsing from images and human inspection and correction. 
Table~\ref{table:dataset} shows a thorough comparison; PubLayNet and PRImA represent two typical types of existing layout analysis datasets: large and automatically-generated v.s. small and human-labeled, with HJDataset an intermediate case.

\textbf{Labeling Simulation} 
When running simulations, we build two additional helper algorithms to imitate human labeling behavior.
First, for the selected objects, the corresponding ground-truths is found via a best-matching algorithm.  
For each prediction, we calculate the IOU with all ground-truth objects and choose the top one to substitute the prediction. 
Duplicated ground-truths selected in an image will be removed by this process. 
In real-world labeling experiments, we also notice human annotators do not need to correct an object prediction if it is accurate (high IOU with the ground-truth and category is the same), which further reduces labeling expense. 
To best simulate this phenomena, if a selected prediction has an IOU$>$0.925 (determined empirically) with some ground-truth objects of the same category, we do not substitute it with the ground-truth and only use a discounted budget $\eta=0.2$.
Finally, to mimic annotators' search for false-negative regions, we compute the pairwise IOU between the ground truth $Y_i$ and the combined labeling objects $\tilde{Y_i}$. 
Ground-truth objects whose maximum IOU with predicted objects is less than $\zeta$ are chosen to add to $\tilde{Y_i}$, and the remaining budget is reduced accordingly.  
$\zeta$ is set to 0.05 in the following experiments to allow minor overlapping caused by noise in the predictions. 

\textbf{Implementation} 
The proposed algorithms are implemented based on Detectron2~\cite{wu2019detectron2}, an open-source object detection benchmark.
For fair comparison, the same object detection model (Faster R-CNN~\cite{ren2015faster} with ResNet-50~\cite{he2016deep} backbone and FPN~\cite{lin2017feature}) is used for all the experiments. 
The optimizer is based on SGD with Momentum~\cite{sutskever2013importance} and MultiStep learning rate warmup~\cite{goyal2017accurate} of 0.00025 base learning rate). 
We train each model on a single Tesla V100 GPU with a batch size of 6. 

The total labeling budget $m$ and the total round $T$ are set per dataset to account for different dataset sizes, and the labeling budget is evenly distributed for each round. 
For the object selection ratio, by default, we use a linear decay function with a given initial and last value. 
These hyperparameters are initialized as indicated in Table~\ref{table:dataset}. 
When calculating the object scores, we set $\lambda$ to 1 and $L$ as the cross entropy function. 
In addition, unless otherwise mentioned, we use four pairs of $(\alpha,\beta)$'s: $(0.08,0.04)$, $(0.08,0.16)$, $(0.12, 0.04)$, $(0.12, 0,16)$, and for each pair, four boxes are created (moving towards top left, top right, bottom left, and bottom right).
A total of $K=16$ perturbed boxes are generated per object prediction for comprehensive analysis of prediction performance under small and large perturbations in different directions.

\section{Results and Discussion}
\label{sec:res}

\subsection{Better AP with the Same Budget}
\label{sec:exp-same-object}

\Olala based labeling settings are compared against image-level AL and other random labeling baselines:
\begin{itemize}
    \item[{[}a{]}] Image-Random: randomly select Images in each round
    \item[{[}b{]}] Image-Marginal: image-level Active Learning baselines~\cite{brust2018active} with marginal scoring and mean aggregation
    \item[{[}c{]}] \Olala-Random: randomly select Objects in each round
    \item[{[}d{]}] \Olala-Marginal: select objects using marginal scoring for object category prediction
    \item[{[}e{]}] \Olala-Perturbation: select objects using the proposed perturbation-based scoring function (Section~\ref{sec:scoring})
\end{itemize}
Given the same (object) labeling budget, we compare the final trained model accuracy among these models. 
According to Table~\ref{table:performance-object-fixed}, object-level annotations, even \Olala-Random cases, are generally better than their image-level counterparts.
\Olala labeling settings usually lead to 1 or 2 points of performance improvements compared to the Image-level AL baselines. 
And the perturbation-based object scoring behaves more robustly as opposed to the marginal scoring method, which only considers object category predictions. 
It is also worth noticing \Olala could annotate significantly more images (Labeled $I$ in the table), especially for larger-scale labeling tasks: in PubLayNet experiments, the created dataset size is 80\% larger (3686 vs. 2046). 

\subsection{Similar AP with Less Budgets}
\label{sec:exp-same-image}

Table~\ref{table:performance-image-fixed} shows the final AP of trained models when labeling the same amount of images. 
With only a part of objects requiring annotation per image, \Olala-based methods considerably reduce the object budget expense.
We observe at most 50\% reduction in the number of labeled objects compared to random image labeling cases in PubLayNet experiments (7496 vs. 15980). 
Moreover, with this level of reduction, \Olala-based models manage to maintain a comparable level of accuracy. 
Similarly, the marginal scoring baseline is less stable and the performance is worse compared to the perturbation-based scoring method in \Olala settings. 

In Figure~\ref{fig:publaynet-image-fixed}, we visualize the model validation accuracy (line plot) and the budget expense (bar plot) for PubLayNet dataset labeling simulations. 
\Olala-based methods expend far less object labeling budget. 
Given the same number of object budget (dashed horizontal line), image-AL methods can only label 5 rounds, and the model AP is around 45 (indicated by the vertical line) and significantly lower than 58.9 in \Olala models. 

\subsection{Analysis of the \Olala framework}
\label{sec:exp-olala}

\begin{figure}[t]
    \centering
    \includegraphics[width=\linewidth]{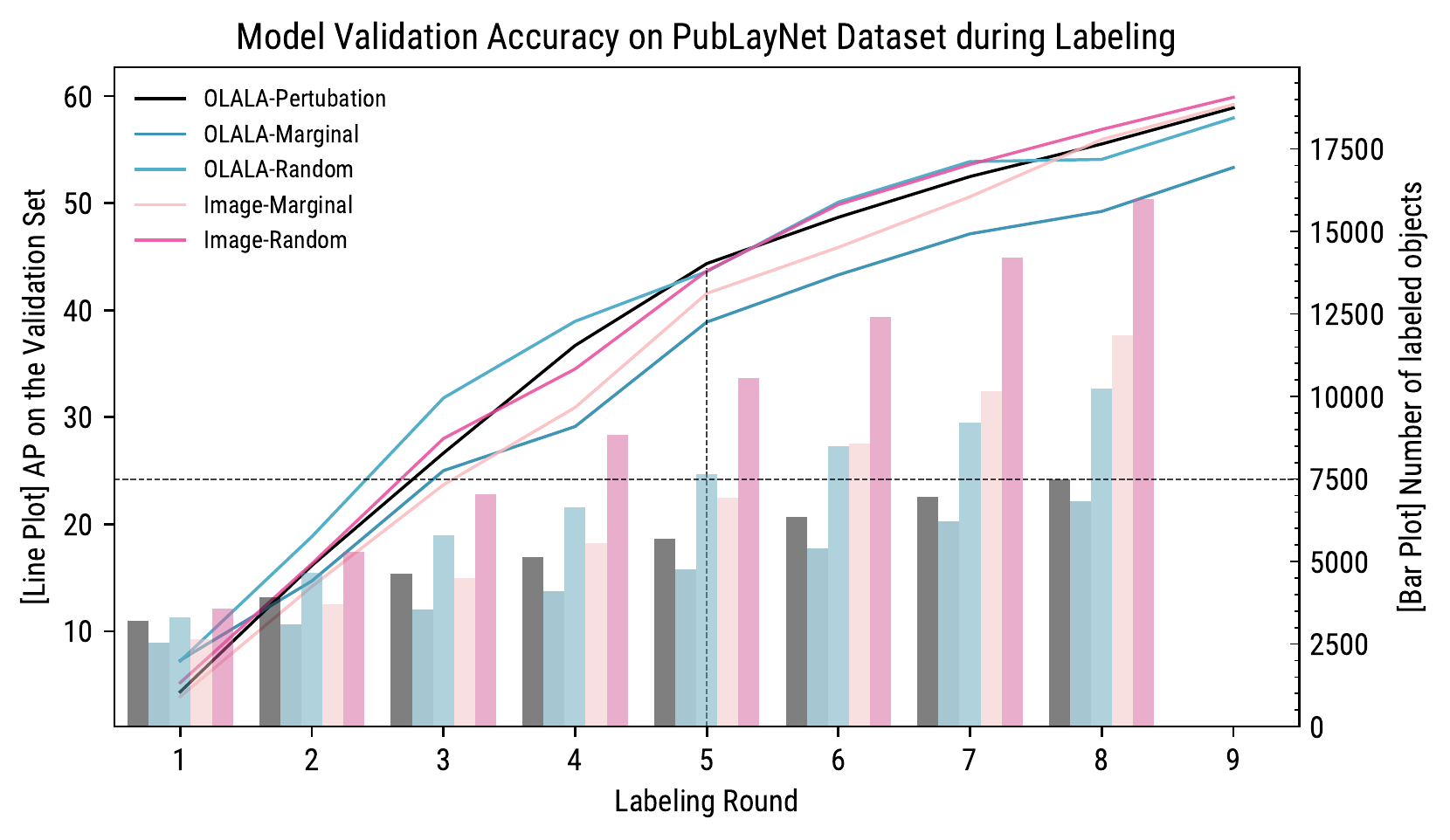}
    \caption{Model validation accuracy (line plot) and budget expenses (bar plot) at different rounds of in PubLayNet labeling. \Olala methods (blue) require labeling less objects compared to image AL methods (red), while maintaining similar AP. If the same number of objects is allowed (horizontal dashed line), image AL method stops at round 5, and the model AP will be around 25\% lower compared to \Olala settings.} 
    \label{fig:publaynet-image-fixed}
    \vspace{-4mm}
  \end{figure}

\begin{figure}[t]
    \centering
    \includegraphics[width=\linewidth]{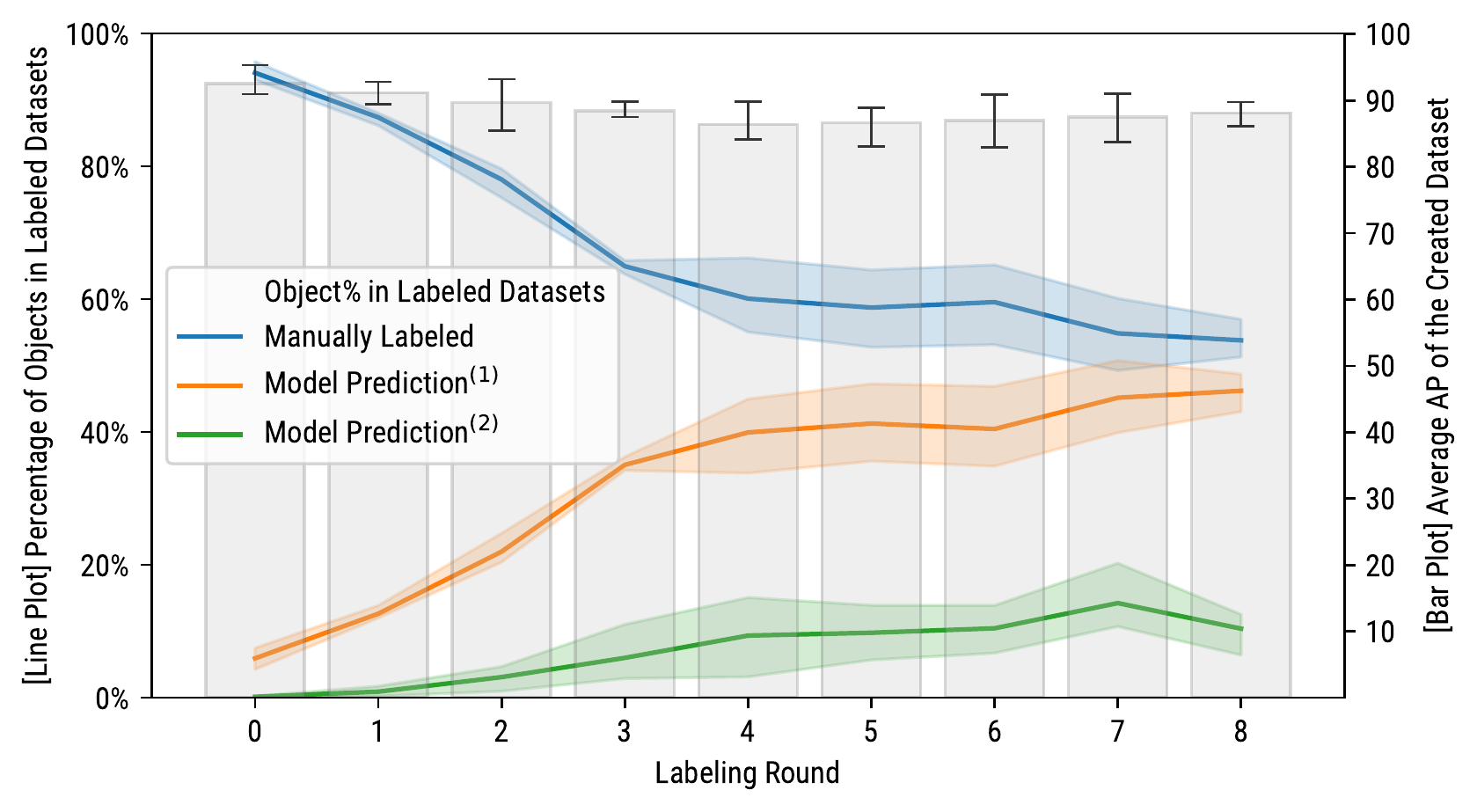}
    \caption{The created object sources (line plot) and dataset accuracy (bar plot) during the training process. The number of manually labeled objects decreases (blue) and directly used model predicted objects (1) portion increases (orange). As model becomes more accurate, higher portion of selected objects (2) become accurate and also unchanged (green). Results shown are averaged from the three \Olala methods in PubLayNet experiments.} 
    \label{fig:labeling-stats}
    % \vspace{-3mm}
  \end{figure}

In the \Olala framework, there are three sources of objects in the created dataset, namely, human annotations, directly used model predictions (unselected in the AL step), and unchanged model predictions (they are selected for manual check, but remain unchanged as they are accurate). 
\Olala strategically choose objects to label and thus optimize the overall efficiency. 
Figure~\ref{fig:labeling-stats} shows the proportion of object sources in the three \Olala settings in PublayNet Labeling experiments. 
The \emph{Object Selection Scheduling} (Section~\ref{sec:scheduling}) sets high selection ratio $r$ when training begins and $r$ decays during training. 
Thus, the averaged percentage of manually labeled objects (blue line) is initially high but gradually decreases while the portion of model prediction (orange line) steadily grows in the labeling process. 
As models becomes more accurate as training progresses (reflected in Figure~\ref{fig:publaynet-image-fixed}), ``annotators'' find more accurate objects in the model-selected predictions, and include them in the dataset without changing them (green line). 
Though more than 50\% of objects directly comes from model prediction, the created datasets still maintain the same high level of accuracy\footnote{The dataset accuracy is measured in AP via comparing the created version with the oracle, in the same style of evaluating model predictions.}, indicated by grey bar plots in the background. 

\begin{table}[t]
    \centering
    \resizebox{\linewidth}{!}{
    \begin{threeparttable}
        \begin{tabular}{r|cc|cc}
        \toprule
        Datasets           & \multicolumn{2}{c|}{PubLayNet}     & \multicolumn{2}{c}{HJData}        \\
        \midrule
        Exps\tnote{1} & AP                    & Labeled $I$/$O$  & AP                    & Labeled $I$/$O$ \\
        \midrule
        {[}a{]}\tnote{2}   & 59.89                 & 1503/15980 & 63.42                 & 603/44156 \\
        {[}c{]}   & 57.96(-1.93)          & 1503/10228 & 65.72(+2.30)          & 603/29191 \\
        \midrule
        {[}b{]}   & 59.21                 & 1503/11848 & 69.04                 & 603/44251 \\
        {[}d{]}   & 53.33(-5.88)          & 1503/6829  & 65.84(-3.19)          & 603/30251 \\
        {[}e{]}   & \textbf{58.90(-0.31)} & 1503/\textbf{7496}  & \textbf{67.68(-1.36)} & 603/\textbf{28899} \\
        \bottomrule
    \end{tabular}
    \begin{tablenotes}
        \small
        \item[1] The parameters in these experiments are slightly different from those mentioned in Table~\ref{table:dataset}, and we will report the details in the supplementary materials.  
        \item[2] The indexing is the same as Table~\ref{table:performance-object-fixed}.
    \end{tablenotes}
    \end{threeparttable}
    }
    \caption{The final AP and number of total labeled images $I$ and objects $O$ when labeling the same number of images. \Olala maintains similar level of AP while requires labeling significantly less number of objects. Similar results are observed in PRImA and abbreviated to save space.}
    \label{table:performance-image-fixed}
    \vspace{-3mm}
\end{table}

\begin{figure}[t]
    \centering
    \includegraphics[width=0.985\linewidth]{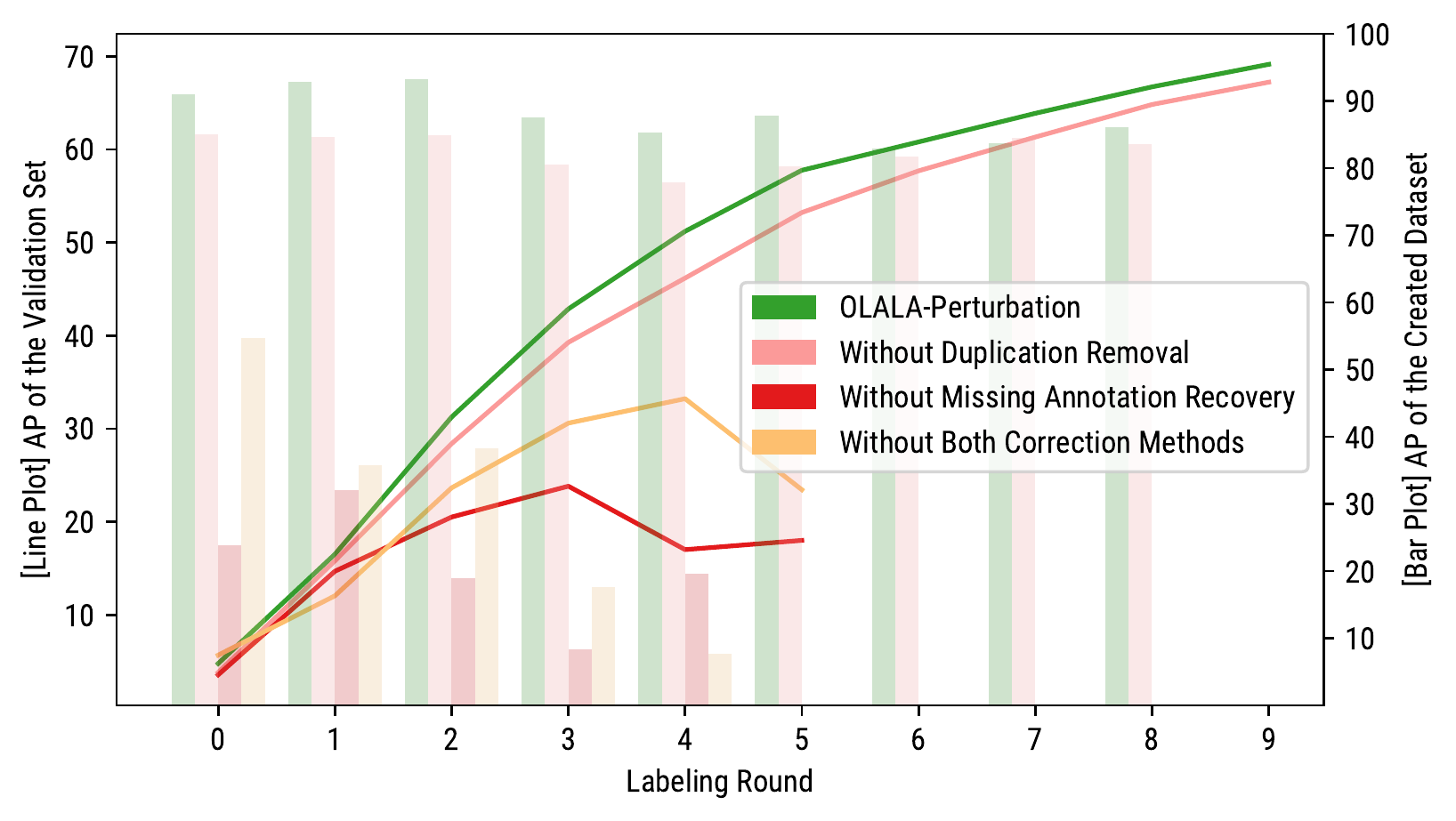}
    \caption{Influence of the prediction correction components on the model validation accuracy (line plot) and dataset accuracy (bar plot). Model performance suffers from removing the components, and dataset accuracy decreases accordingly. The \emph{Missing Annotation Recovery} component is critical to model performance by correcting false negatives from predictions. Results shown are from experiments on PubLayNet.} 
    \label{fig:error-fixing}
    % \vspace{-3mm}
  \end{figure}

On the other hand, we study how the semi-automatic prediction correction algorithm, mentioned in Section~\ref{sec:correction}, contributes to the \Olala process. 
Shown in Figure~\ref{fig:error-fixing}, we compare the model validation AP (line plot) and accuracy of the created dataset (bar plot) with and without the \emph{Duplication Removal} and \emph{Missing Annotation Recovery} components in PubLayNet annotation. 
Without these components, models suffer from different levels of accuracy reduction compared to the \Olala-Perturbation baseline (green), and the dataset. 
We observe the most severe accuracy reduction when removing the missing annotation recovery components (red), indicating the necessity of extra supervision for correcting high ratios of false negatives. 
Interestingly, when removing both correction method (orange), the model appears to perform better than only discarding missing annotation recovery component. 
However, they are not contradictory.
We find that duplicated predictions add more instances per image for calculating the loss, thus reinforce the signal to train the model and improves the initial performance. 
Unfortunately, without extra supervision, the models are trained on a dataset with many false negatives, and tend to generate less predictions.
The error accumulates and finally both models collapse and stop improving. 
And they exhaust all the training samples at round 5\footnote{In this case, \Olala only selects the top $r$ percentage of objects for human correction and counts for budgets for these objects. Therefore, as these models generate few predictions per page, they expend less budget per image and quickly ``label'' all images in the dataset, though incorrectly.}. 

\section{Related Work}
\label{sec:related}

\textbf{Document Layout Detection} poses significant challenges because of the complex organization of layout objects and many objects
types that may be present~\cite{lee2013pseudo, clausner2019icdar2019, gao2017icdar2017, antonacopoulos2015icdar2015}. 
In order to train layout detection models, researchers have created various datasets for historical manuscripts~\cite{simistira2016diva, gruning2018read}, newspapers~\cite{clausner2015enp}, and modern magazines~\cite{antonacopoulos2009realistic}. 
Due to the prohibitive labeling cost for annotating for many layout objects per page, they typically contain only hundreds of labeled pages, and are not sufficient to train and evaluate deep learning based models (e.g., Faster R-CNN~\cite{ren2015faster}). 
Recently, there have been efforts to generate large-scale document dataset automatically via parsing electronic PDF documents~\cite{zhong2019publaynet, zhong2019image}. 
Unfortunately, this approach is not generalizable, \eg, to millions of scanned document images, which has the potential to make fundamental contributions to important research questions in business, the social sciences, and the humanities. 

\textbf{Active Learning} has long been applied to object detection. 
Abramson and Freund~\cite{abramson2006active} use AL for efficiently labeling pedestrians by sampling images based on predictions generated by an AdaBoost model. 
Yao \etal~\cite{yao2012interactive} study an annotator-centered labeling cost estimation method and prioritize labeling for high-cost images. 
In the context of deep learning, image level scores are generated via aggregation of marginal scores for candidate boxes~\cite{brust2018active} or applying query by committee~\cite{seung1992query} to features maps~\cite{roy2018deep}.
Aghdam \etal~\cite{aghdam2019active} propose a pixel level scoring method using convolution backbones and aggregate them to informativeness scores for image ranking. 
In general, most related AL works concentrate on image-level scoring and selection. 

\textbf{Self-training} techniques~\cite{rosenberg2005semi} has seen applications in semi-supervised learning settings.
Recent work~\cite{wang2018towards, xie2020self} has demonstrated that predictions on unlabeled images could boost model performance.
Desai \etal~\cite{desai2019adaptive} also report that weak labels can be used to improve model predictions during training. 
Yet without guarantees of their accuracy, the model generated (pseudo) labels on unlabeled data are usually disposed after the training loop, and wrongful prediction might even hurt the model training. 
Therefore, self-training is usually considered as a technique for improving model performance but does not help create a larger dataset.

\section{Conclusion}

In this paper, we propose the object-level active learning annotation framework, \Olala, for efficiently labeling layout images. 
With a novel prediction correction algorithm and perturbation object scoring function, annotators only need to label a part of layout objects in each image.
Through simulated labeling experiments on real-world data, we show that our proposed algorithms significantly improve dataset creation efficiency relative to image-level methods. 
Different components of \Olala are also carefully studied to demonstrate their validity and necessity. 
The remarkable efficiency gain of \Olala can benefit many downstream tasks, such as the processing of historical documents at scale or annotating large bulk of financial forms. 
To sum up, our work explores better cooperation between human and machine intelligence, and have the potential to unlock novel data annotation paradigms in general. 

{\small
\bibliographystyle{ieee_fullname}
% Original ieee_fullname does not support natbib \citet command. I found the script from https://gist.github.com/Randl/bbdf85e7028716f39a69d7828d07ef33, which helps enables this functionality. 

\bibliography{egbib}
}

\clearpage
\appendix

\twocolumn[{%
\renewcommand\twocolumn[1][]{#1}%
\maketitle
\begin{center}
    \centering
    \includegraphics[width=\textwidth]{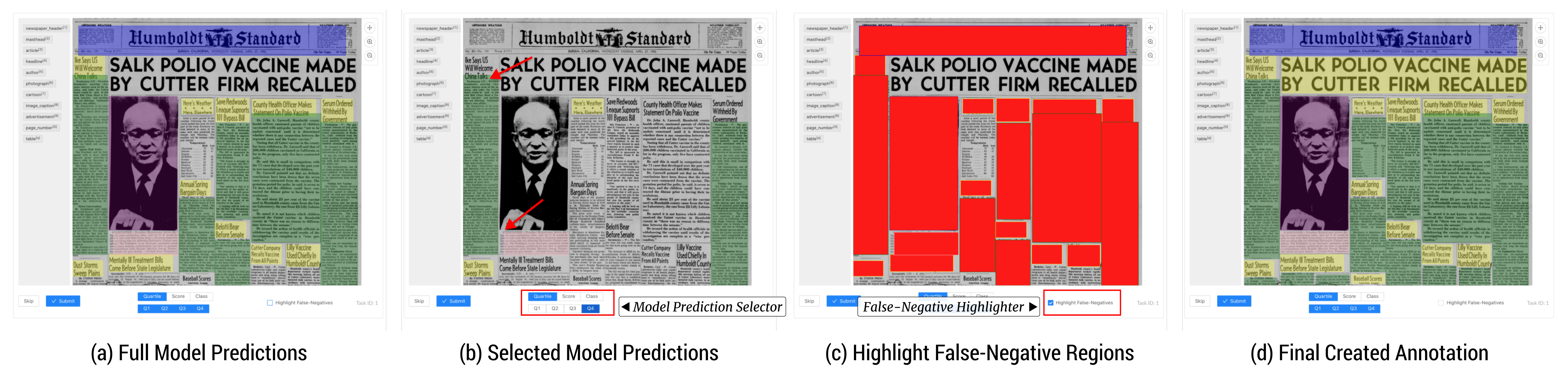}
    \captionof{figure}{Illustration of the annotation interface with \Olala features. 
    (a) Given an input scan, a pre-trained model generates object predictions, and they are highlighted as rectangular boxes on the original image. The color denotes the category of the given object.
    (b) The \emph{Model Prediction Selector} enables hiding predictions of low object scores. In this case, objects of top 25\% (the 4th Quartile, Q4) scores are presented. Two of the objects (pointed by red arrows) have minor errors in object location predictions. Human annotators check only the displayed objects and modify inaccuracies. 
    (c) The \emph{False-Negative Highlighter} helps recognize mis-identified objects from the model predictions. After enabled, it converts all predicted regions to a dummy color, and regions without predictions are highlighted. Annotators can easily spot false-negatives regions and have them labeled.
    (d) After these steps, the full image annotation is created with less effort. 
    }
    \label{fig:ui} 
    \vspace{1mm}
\end{center}%
}]
\section*{Appendix}

\section{\Olala Implementation Details}
Different from image-level labeling, annotating objects within images is fundamentally a search task: ``annotators''\footnote{We use the general term annotator to refer to a human annotator or a simulated labeling agent.} need to scan through the image and find objects matching specific criteria. 
The nature of object-based labeling leads to different objectives in simulated labeling experiments and real-world human annotation.
In labeling simulations, the ground-truth objects are known ex-ante. 
The labeling agent only needs to query the oracle and choose objects that meet certain conditions. 
As the search space is pre-defined, the core challenge is to construct such query conditions for finding ground-truths.
By contrast, when humans annotate objects, there is no ground-truth known beforehand, and the object search space is yet undefined. 
Their vision systems are capable of efficiently identifying correct objects within the space. 
Hence, the objective for human annotation is to reduce the object search space, and annotators will select valid objects within the space. 
To this end, as mentioned in the main paper, the \Olala framework is implemented differently for real-world human annotation (Section~\ref{sec:ui}) and simulated labeling experiments (See Section 4 in the main paper).% (Section~\ref{sec:simulated-agent}).

\begin{figure*}[t]
    \centering
    \includegraphics[width=\linewidth]{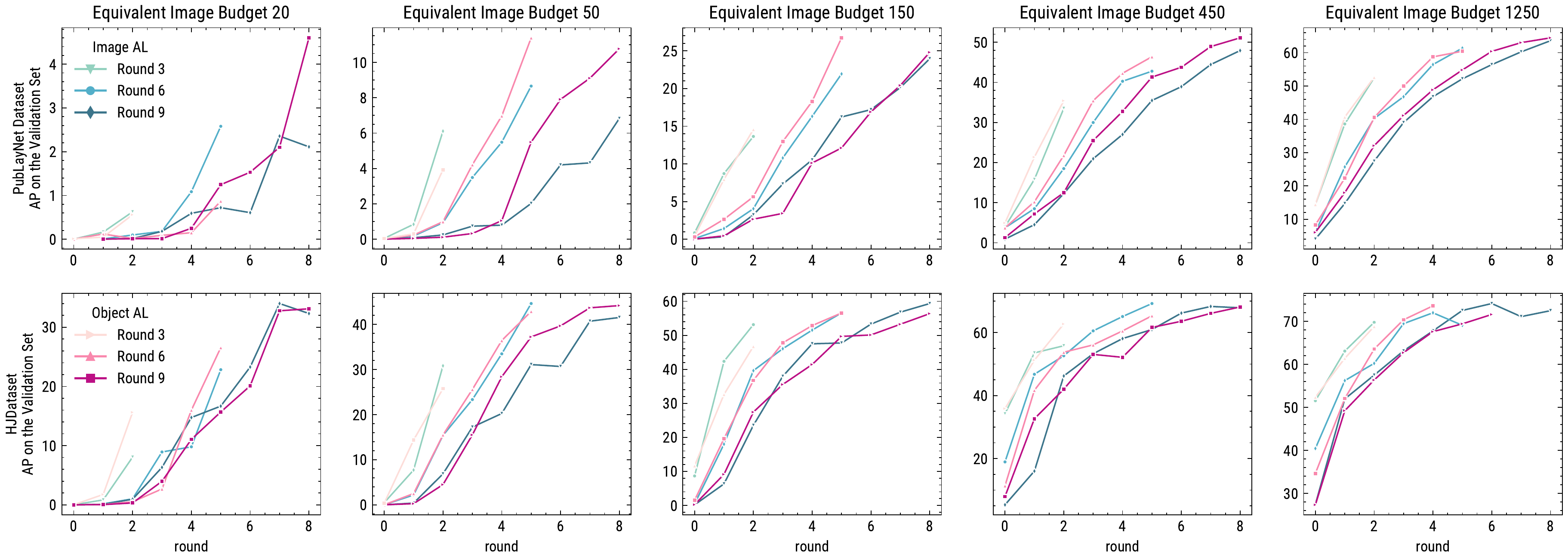}
    \caption{The model validation AP during the labeling process under different total rounds $T$ and labeling budget $m$. The plots in row one and two are for experiments on the PubLayNet and HJDataset, respectively. Within each plot, image-level AL results are colored in blue while \Olala results are in red. To best show results at different stage of training, the ranges for y-axis are set differently. Under the same budget, increase $T$ can generally lead to better model performance. For different datasets, the optimal budget and total round settings are different. As the number of budget increases, image-level methods narrow the performance gap (in PubLayNet experiments) or perform better than \Olala methods (in HJDataset experiments).} 
    \label{fig:compare-m-t}
\end{figure*}

\subsection{\Olala Annotation User Interface}
\label{sec:ui}

To help with human annotation, we build a labeling interface incorporated with \Olala functionalities based on label-studio\cite{Yuxin2020label}. 
Figure~\ref{fig:ui} shows an example of annotating newspaper layouts using this tool\footnote{In this example, the used model has been trained on 200 hundred images. For illustration purpose, we reduce the number of objects generated by models to emphasize the false-negative selection process. But in practice, the false-negative rate is lower.}. 
\begin{enumerate}
    \item[a] Given an input scan, a pre-trained model generates object predictions $\{(b_j, c_j)\}_{j=1}^n$, which are highlighted as rectangular boxes on the original image. 
    The color denotes the category $c_j$ of an object.
    Within the outputs, duplicated object detections are precluded using \emph{Duplication Removal}. 
    \item[b] A \emph{Model Prediction Selector} is implemented for hiding objects with low scores generated by the object scoring function $f$. 
    In this case, objects of top 25\% (the 4th Quartile, Q4) scores are presented.
    Two selected objects (pointed by red arrows) have minor errors in object location predictions by missing one line or one column of text (see Section 3 ``Applicability to Layout Datasets'' in the main paper), while others being correct.
    Human annotators can focus on checking the displayed objects and only need to modify the two incorrect predictions while other accurate ones are kept untouched. 
    \item[c] We also develop a \emph{False-Negative Highlighter} to help annotators find mis-identified objects from the model predictions. 
    After enabled, it will assign a dummy color overlay to object predictions, thus regions without predictions will be highlighted. 
    Annotators can easily spot false-negatives regions and have them labeled. 
    And this is the \emph{Missing Annotation Recovery} step in the \Olala algorithm. 
    \item[d] Finally, the full image annotation will be created with significantly less effort. 
\end{enumerate}

Through the interface, annotators' labeling effort is saved via a reduced object search space: one only needs to check the selected model predictions and the highlighted false-negative regions. 

\begin{table}[t]
    \resizebox{1.\linewidth}{!}{
    \begin{threeparttable}
        \begin{tabular}{r|cc|cc}
            \toprule
            \textbf{Configuration}   & \multicolumn{2}{c|}{\textbf{Configuration A}} & \multicolumn{2}{c}{\textbf{Configuration B}} \\ 
            Datasets            & PubLayNet           & HJDataset           & PubLayNet           & HJDataset           \\
            \midrule
            Labeling budget $m$ & 21,140              & 51,436           & 15,855              & 44,088           \\
            Equivalent image budget   & 2,000\tnote{1}      & 700              & 1,500               & 600              \\
            Total rounds $T$    & 10                  & 8                & 9                   & 9                \\
            Initial / last $r$  & 0.9/0.4             & 0.9/0.5          & 0.9/0.5             & 0.9/0.5          \\
            \bottomrule
        \end{tabular}
        \begin{tablenotes}
            \small
            \item[1] To get the number of equivalent image budget, we simply divide $m$ by the average number of objects per page for the given dataset. 
        \end{tablenotes}
    \end{threeparttable}
    }
\caption{Different parameter configurations for labeling settings (1) and (2). Configuration A is used for labeling setting (1) where the same number of objects are labeled; and B for labeling setting (2) where the number of labeled images is fixed. }
\label{table:two-para-configs}
\end{table}
\begin{table}[t]
    \centering
    \resizebox{\linewidth}{!}{
    \begin{threeparttable}
        \begin{tabular}{r|cc|cc}
        \toprule
        Datasets  & \multicolumn{2}{c|}{PubLayNet}     & \multicolumn{2}{c}{HJData}        \\
        \midrule
        Exps      & AP                    & Labeled $I$/$O$  & AP        & Labeled $I$/$O$ \\
        \midrule
        {[}a{]}   & 61.65                 & 1558/16123   & 62.73         & 605/44505  \\
        {[}c{]}   & 63.73(+2.07)          & 2501/16122   & 65.75(+3.02)  & 980/44260  \\
        \midrule
        {[}b{]}   & 65.52                 & 1961/16108   & 68.16         & 607/44344  \\
        {[}d{]}   & 69.36(+3.83)          & 2995/16104   & 69.13(+0.97)  & 956/44398  \\
        {[}e{]}   & 65.53(+0.01)          & 2996/16142   & 69.15(+0.99)  & 1041/44398 \\
        \bottomrule
    \end{tabular}
    \end{threeparttable}
    }
    \caption{The final AP and number of total labeled images $I$ and objects $O$ when labeling the same number of objects under model configuration B.}
    \label{table:config-B-object}
    \vspace{-2mm}
\end{table}

\section{Additional Experiments}

\subsection{Different model configurations}

In the main paper, we report results under two different settings, namely, (1) labeling the same number of objects and (2) labeling the same number of images. 
During these experiments, the model configurations for labeling settings (2) is slightly different than those in (1), and we include the details in Table~\ref{table:two-para-configs}. 
Labeling setting (1) is only experimented under configuration A while (2) under configuration B. 
For fair comparison, we complete another set of experiments for labeling setting (1) using configuration B. 
The results are reported in Table~\ref{table:config-B-object}, and similar conclusion could be made based on this set of experiments.

\subsection{Analysis of labeling budget and total training rounds}

We run additional labeling simulations to find the optimal configurations for the labeling budget and the total training rounds. 
Given the same budget, we could perform multiple rounds of labeling and re-training, with the optimal total round yet to be determined. 
Similarly, for a given dataset, it is important to allocate appropriate labeling budget such that the labeled samples can most effectively boost the model performance. 
This study could also shed light on the applicability of \Olala to labeling scenarios where only small labeling budget is allowed.
To this end, we experiment with object budget $m$ equivalent to labeling 20, 50, 150, 450, and 1250 images (equivalent image budget \footnote{Directly setting thresholds for $m$ does not account for the variances of objects per image for different datasets.}) for a given dataset. 
For each $m$, we also experiment with three different total labeling rounds $T$ of 3, 6, and 9. 
The model validation accuracy during the labeling process is visualized in Figure~\ref{fig:compare-m-t}.

Given the same labeling budget, we find that increasing the total labeling rounds $T$ tends to improve the model accuracy, especially for scenarios where small labeling budget is available. 
Under such small budget, \Olala-based annotation usually leads to models of higher accuracy than those from image-level AL settings. 
However, as labeling budget increases, the performance gap between \Olala and image AL models narrows. 
With sufficient labeling budget, image AL models even performs better than \Olala models in HJDataset. 
It reveals that \Olala is more helpful in the initial stage of labeling, as it exposes more images samples to the model and thus boosts the performance.
For different datasets, the optimal combination of total labeling rounds and budget is different: $T=9$ with the equivalent image budget of 450 for PubLayNet, and $T=9$ with 50 equivalent image budget for HJDataset. 
Based on our observation, this is largely determined by the diversity of samples in the dataset. 
\Olala helps to explore unique object instances in the early training stage, and requires more labeling steps to achieve optimal performance boost for datasets of diverse examples like PubLayNet.

\end{document}